\documentclass[10pt,letterpaper,twocolumn]{article}

\usepackage[OT1]{fontenc}
\usepackage{times}
\usepackage{amsmath,amssymb}
\usepackage{graphicx}
\usepackage{xcolor}
\usepackage{booktabs,tabularx,listings}
\usepackage[font=small,labelfont=bf,labelsep=period]{caption}
\usepackage[numbers,sort&compress]{natbib}
\usepackage[hyphens]{url}
\usepackage{flushend}
\usepackage[pdfversion=1.7]{hyperref}

\title{AeroSkill: From Composable Skills to Long-Horizon Aerial Autonomy}
\definecolor{linkblue}{HTML}{234D77}
\hypersetup{
  pdftitle={AeroSkill: From Composable Skills to Long-Horizon Aerial Autonomy},
  pdfauthor={Wenxiang Huo, Kaibo Yang, Enze Zhu, Shijuexi Wang, Xuyang Li, Yajie Chai, Ran Song, Chengxin Liu, Xiaoxuan Liu, Lei Wang, Zhan Chen},
  colorlinks=true,
  allcolors=linkblue
}

\newcommand{\affilmark}[1]{\textsuperscript{#1}}

\author{%
  Wenxiang Huo\affilmark{1}, Kaibo Yang\affilmark{1}, Enze Zhu\affilmark{1},
  Shijuexi Wang\affilmark{2}, Xuyang Li\affilmark{1}, Yajie Chai\affilmark{4}\\[3pt]
  Ran Song\affilmark{3}, Chengxin Liu\affilmark{3}, Xiaoxuan Liu\affilmark{1},
  Lei Wang\affilmark{1}, Zhan Chen\affilmark{1}}
\date{}

\makeatletter
\renewcommand{\section}{%
  \@startsection{section}{1}{\z@}{-10pt plus -2pt minus -2pt}{7pt}{\normalfont\large\bfseries}}
\renewcommand{\subsection}{%
  \@startsection{subsection}{2}{\z@}{-8pt plus -2pt minus -2pt}{5pt}{\normalfont\fontsize{10.25}{12.4}\selectfont\bfseries}}
\renewcommand{\subsubsection}{%
  \@startsection{subsubsection}{3}{\z@}{-6pt plus -2pt minus -2pt}{3pt}{\normalfont\normalsize\bfseries}}
\renewcommand{\paragraph}{%
  \@startsection{paragraph}{4}{\z@}{5pt plus 2pt minus 1pt}{-0.7em}{\normalfont\normalsize\bfseries}}
\renewcommand{\@seccntformat}[1]{\csname the#1\endcsname.\enspace}

\renewcommand{\@maketitle}{%
  \begin{center}
    {\fontsize{16}{19}\selectfont\bfseries\@title\par}
    \vspace{10pt}
    {\large\@author\par}
    \vspace{6pt}
    {\normalsize
      \affilmark{1}\,University of the Chinese Academy of Sciences
      \qquad \affilmark{2}\,Tianjin University\\[2pt]
      \affilmark{3}\,Shandong University
      \qquad \affilmark{4}\,Baidu Inc.\par}
  \end{center}%
}
\makeatother

\begin{document}
\maketitle

\begin{abstract}
Complex language-guided tasks often require unmanned aerial vehicles (UAVs) to execute a sequence of flight operations. Each operation has different motion requirements and spatial constraints. Environmental disturbances and accumulated errors can affect individual operations, while state mismatches between operations can compromise subsequent control. To address these challenges, we present AeroSkill, a hierarchical execution framework based on parameterized flight skills. A unified skill representation connects task decomposition, action correction, and progress estimation. The framework converts instructions into skill sequences and uses a lightweight residual adapter to correct base actions during execution. The separate progress estimator guides this residual correction and determines when each skill ends. In addition, we introduce a visually grounded transition policy conditioned on the requirements of the next skill. It uses visual feedback to establish suitable starting states for subsequent skills. This improves the reliability of the transition and supports the continuous execution of complex long-horizon tasks. The framework organizes the pretrained model's control capabilities into reusable skills. By composing these skills, it supports the execution of diverse long-horizon tasks. We evaluate the framework on UAV-Flow and FLIGHT, where it improves the success rate over the baseline by 19.05 points and 11.50 points, respectively. Real-world flights further demonstrate AeroSkill's capability for long-horizon tasks.
\end{abstract}
\section{Introduction}
\label{sec:intro}

\begin{figure}[t]
    \centering
    \includegraphics[width=\linewidth]{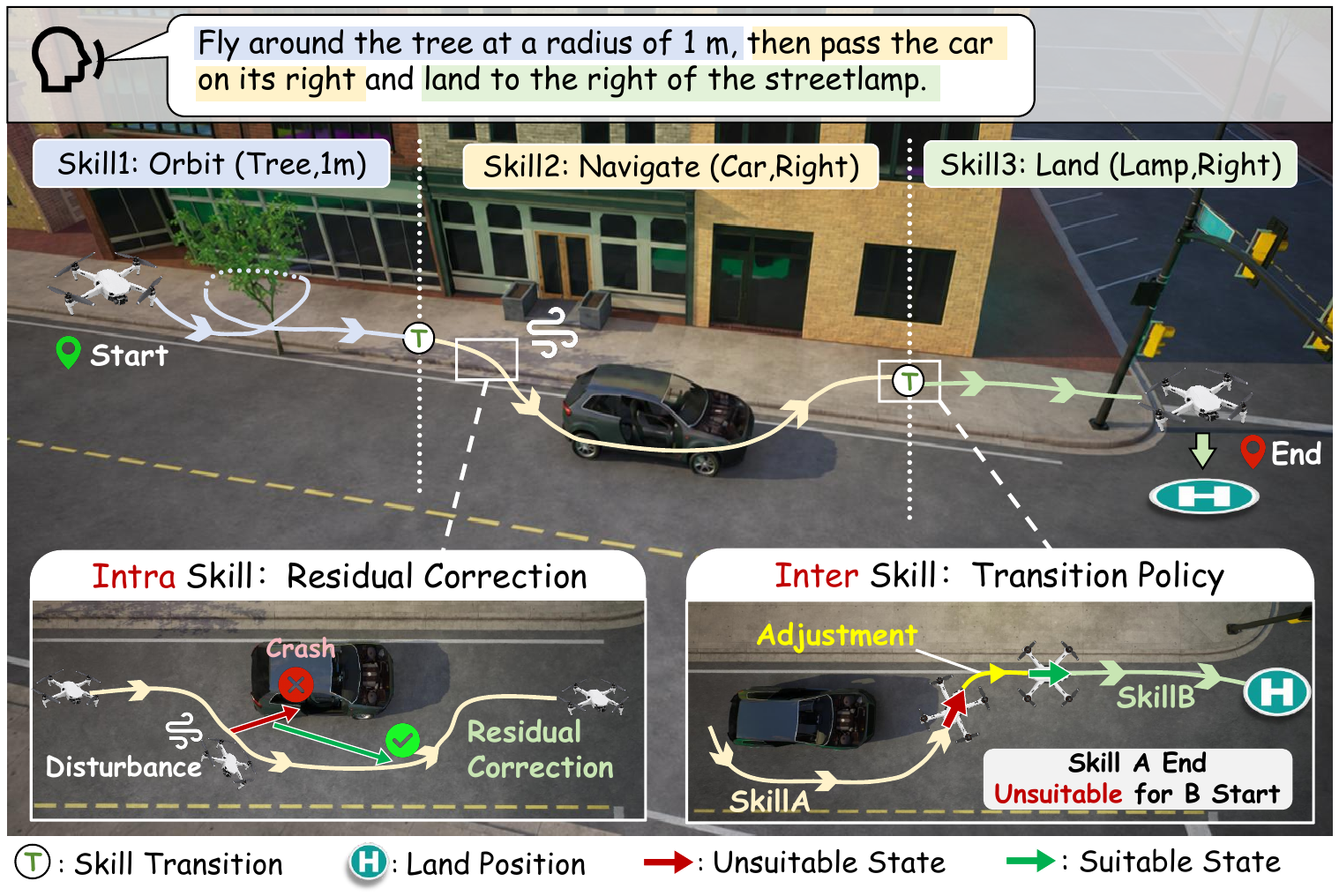}
    \caption{\textbf{Motivation for AeroSkill.}
    (Top) One instruction specifies a sequence of flight skills.
    (Bottom left) Residual correction compensates for disturbances during a skill.
    (Bottom right) The transition policy adjusts the UAV state before the next skill begins.}
    \label{fig:motivation}
\end{figure}

Complex language-guided UAV tasks require translating instructions into precise flight actions.
These actions must satisfy environmental constraints and meet the motion requirements of each stage.
For example, ``Fly around the tree at a radius of 1\,m, pass the car on its right, and land to the right of the streetlamp.''
As the UAV moves through this sequence, it must adapt its motion to different targets and spatial constraints.
The state reached after each operation also determines how the next operation begins.
Reliable execution therefore requires both accurate control within each operation and suitable transitions between operations (Fig.~\ref{fig:motivation}).

Research on UAV vision-language navigation (UAV-VLN) focuses primarily on route following, target localization, and navigation decisions.
UAV vision-language-action (UAV-VLA) research has further explored precise flight control.
For example, UAV-Flow~\cite{wang2025uavflow} learns continuous actions from atomic instructions and expert trajectories.
FLIGHT VLA~\cite{zheng2026flight} uses an asynchronous architecture to coordinate reasoning about task states with high-frequency action generation.
FreeFly-Thinking~\cite{zhou2026freeflythinking} jointly predicts explicit reasoning text and continuous waypoints, connecting semantic decisions with motion prediction.
These advances support action generation for individual operations and long-horizon tasks.
However, their execution remains vulnerable to motion errors and environmental disturbances, while state mismatches between consecutive operations can compromise transition reliability.

These limitations reflect two challenges in executing consecutive flight operations.
First, external disturbances and errors in control response can cause the actual UAV motion to deviate from the requirements of the task.
Uncorrected errors reduce the accuracy of trajectory tracking and final positioning.
Correcting these deviations requires adjusting subsequent actions according to the actual flight state.
The system must also track execution progress to determine whether the current operation is complete.
Second, completing one stage does not necessarily leave the UAV in a suitable state for the next operation.
Different terminal states change the spatial relationship between the UAV and the next target.
Subsequent operations may then begin from unfamiliar states, making the capabilities demonstrated in isolation less reliable.
These two challenges are interrelated: errors within a stage alter its terminal state and the starting conditions of subsequent operations.

To address these challenges, we propose AeroSkill, a hierarchical execution framework based on parameterized flight skills.
It organizes the control capabilities of a pretrained UAV-VLA model into reusable and composable execution units.
Skill types and explicit parameters specify the requirements of each stage, allowing the same flight capabilities to serve different tasks.
To correct motion errors within each stage, a residual adapter uses flight feedback to refine the base actions generated by the frozen UAV-VLA model.
A progress estimator tracks how much of the current operation has been completed.
The residual adapter uses the predicted progress to correct actions according to the current stage of skill execution.
When the predicted progress exceeds a preset threshold, the system ends the current skill and initiates the transition to the next.
By combining skill requirements, action correction, and progress estimation, AeroSkill improves the frozen model's ability to execute consecutive skills accurately and reliably.

To improve transitions between consecutive skills, we introduce a visual transition policy conditioned on the next skill.
It uses the next skill's reference information to attend to relevant features in the current visual observation.
The attended features and skill type guide a single adjustment of the UAV state before control passes to the next skill.
We train skill execution and the transition policy separately, using online interaction for individual skills and offline supervision for transitions.
This approach avoids the need for online interaction over complete task sequences, reducing the overall training overhead.
AeroSkill combines action correction within skills with state adjustments between skills.
This combination supports reliable skill composition throughout the execution of complex, long-horizon tasks.

Our main contributions are as follows.
\begin{enumerate}
    \item
    AeroSkill combines parameterized skills, progress-guided residual control and switching, and visual transitions to enable precise long-horizon execution with a frozen backbone.
    It reaches 79.12\% single-skill SR on UAV-Flow. On FLIGHT, it reaches 70.50\% task SR, 11.50 percentage points higher than FLIGHT VLA.

    \item
    The visual transition policy improves skill composition in long-horizon tasks by preparing the successor's starting state in one step.
    With skill executors unchanged, this transition raises FLIGHT task SR from 64.50\% to 70.50\% and SSR from 73.03\% to 84.22\%.

    \item
    The complete framework transfers to real flights without deployment-specific fine-tuning.
    On the same 100 long-horizon instructions, AeroSkill completes 51/100 tasks, compared with 7/100 for OpenVLA-UAV.
\end{enumerate}
\begin{figure*}[t]
    \centering
    \includegraphics[width=\textwidth]{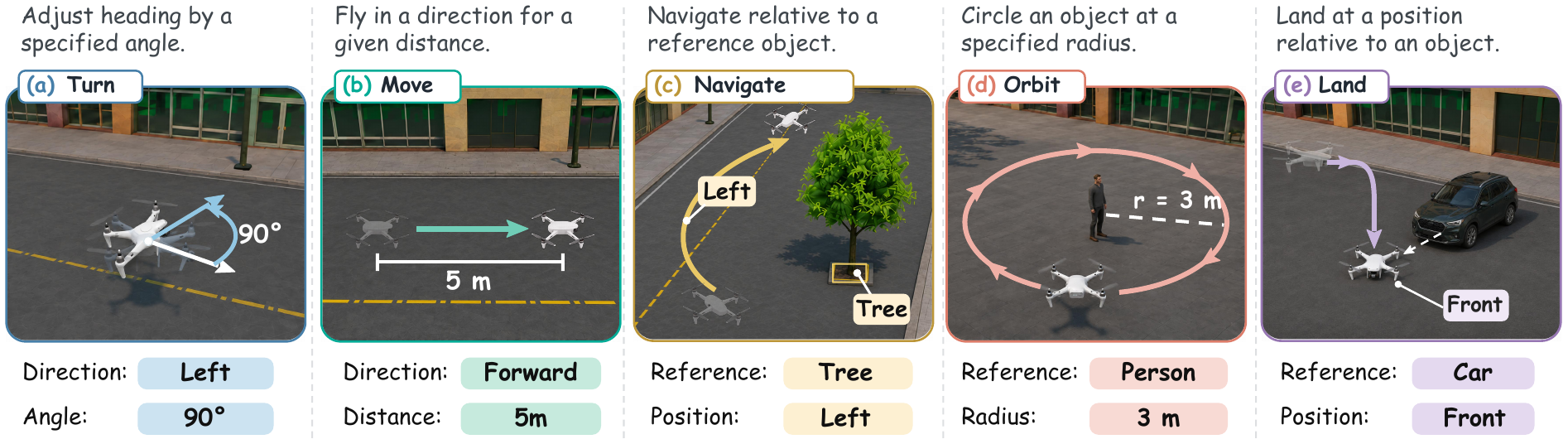}
    \caption{\textbf{Five parameterized flight skills in AeroSkill.}
    Each skill combines a reusable flight behavior with explicit task parameters.}
    \label{fig:flight_skills}
\end{figure*}
\section{Related Work}
\label{sec:related_work}

\subsection{Language-Guided UAV Navigation and Control}
\label{sec:rw_uav_control}

Vision-and-language navigation (VLN) uses visual observations to follow route instructions~\cite{anderson2018vln,krantz2020vlnce}.
NaVid uses video history to choose the next action in continuous navigation~\cite{zhang2024navid}.
AerialVLN and CityNav bring language-guided navigation to outdoor flight~\cite{liu2023aerialvln,lee2025citynav}.
Aerial navigation methods draw on observation history and directional cues~\cite{gao2025openfly,ning2026lookasidevln}, combine imitation and reinforcement learning~\cite{fan2026htnav}, or use dialogue to clarify instructions~\cite{fan2023avdn,chen2026aerialvla}.

Generalist policies such as Octo and OpenVLA learn language-conditioned control from robot demonstrations~\cite{ghosh2024octo,kim2025openvla}.
OpenVLA-OFT studies how action representation and decoding affect control speed and success~\cite{kim2025oft}.
UAV policies address goal-directed navigation, aerial manipulation, and low-level control~\cite{sun2026autofly,sun2026airvla,chen2025gradnav}.
UAV-Flow studies the precise execution of individual maneuvers from short instructions~\cite{wang2025uavflow}.
FLIGHT VLA tackles multi-stage flight by coupling task-state reasoning with high-frequency action generation~\cite{zheng2026flight}.
These studies extend language-guided flight from route following to precise control and multi-stage execution.
However, generating a sequence of actions does not by itself ensure reliable execution within and between flight operations.
AeroSkill addresses these execution requirements through reusable flight skills with action correction and state adjustment between skills.

\subsection{Skill Composition and Execution}
\label{sec:rw_skill_composition}
Skills are commonly modeled as control policies with termination rules~\cite{sutton1999options}.
Skill chaining connects such skills to achieve task goals~\cite{konidaris2009skillchaining}.
Policy sketches and CompILE learn reusable behaviors from task sketches and demonstrations, respectively~\cite{andreas2017sketches,kipf2019compile}.
Language-guided methods organize these behaviors into plans for complex tasks.
SayCan selects skills using affordance scores, while Inner Monologue updates plans using execution feedback~\cite{ichter2023saycan,huang2023innermonologue}.
Code as Policies and ProgPrompt express plans as executable programs~\cite{liang2023codeaspolicies,singh2023progprompt}.
UAV systems similarly compose flight and perception functions to complete tasks~\cite{chen2025typefly,wang2025airstar,li2026aerialclaw}.

Reliable execution additionally depends on action correction and appropriate skill termination.
Residual policies learn corrections to existing controllers~\cite{johannink2019residual,silver2018residual}, including motion primitives and reusable skills~\cite{davchev2022residual,rana2023residualskills}.
Progress estimation supports execution monitoring in VLN~\cite{ma2019selfmonitoring}.
AeroSkill uses the same predicted progress to condition residual correction throughout execution and determine when each parameterized UAV skill ends.
These mechanisms support the continuous execution of instruction-derived skill sequences on a frozen UAV-VLA backbone.

\subsection{Skill Transition Policies}
\label{sec:rw_skill_transitions}
Skill transition methods connect the terminal state of one skill to a suitable starting state for the next skill, supporting execution over longer horizons.
Early work learns transition policies to connect skills, including through state-action distribution matching~\cite{lee2019composing,byun2021transition}.
T-STAR addresses this compatibility during skill learning by regularizing terminal state distributions~\cite{lee2022tstar}.
N2M uses preferences learned from rollouts to select a base pose for subsequent manipulation~\cite{chai2025n2m}.

UAV flight poses additional challenges because of rapid motion and large viewpoint changes.
At skill handoffs, residual motion and targets leaving the field of view can hinder execution of the next skill.
AeroSkill uses the successor's reference information to attend to relevant visual features.
Trained through offline behavior cloning, the policy combines these features with the successor's skill type to adjust handoff states for reliable composition.

\section{Method}
\label{sec:method}

\subsection{Task Formulation and Framework Overview}
\label{sec:method_overview}

We study the precise execution of complex UAV flight tasks guided by natural language.
Given an overall instruction $L$, the UAV must perform consecutive operations while satisfying the motion requirements and spatial constraints of each stage.
At time $t$, the system receives a visual observation $I_t$ and a position state $\mathbf{s}_t$.
It generates an action consisting of relative displacement and a yaw adjustment in the body frame:
\begin{equation}
    a_t=(\Delta x_t,\Delta y_t,\Delta z_t,\Delta\mathrm{yaw}_t).
    \label{eq:action_space}
\end{equation}
Each action lasts one control cycle.
Each operation begins from the preceding terminal state, so the flight sequence depends on precise skill execution and reliable handoffs.

\begin{figure*}[t]
    \centering
    \includegraphics[width=\textwidth]{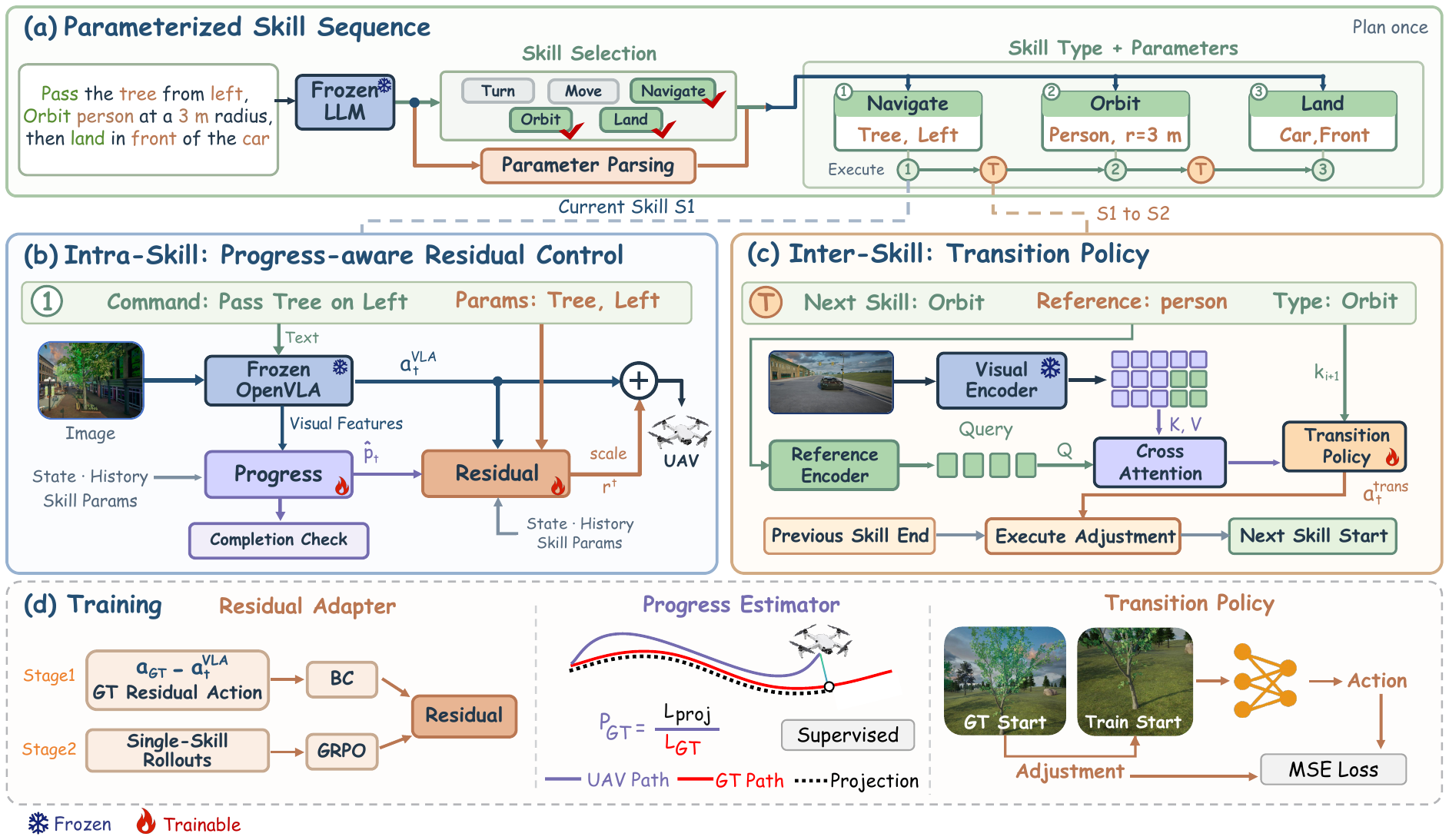}
    \caption{
        \textbf{Overview of AeroSkill.}
        (a) A frozen LLM parses the task instruction into an ordered sequence of parameterized flight skills.
        (b) Skill-specific residual adapters refine actions from a frozen UAV-VLA model, with predicted progress guiding correction and skill switching.
        (c) The next skill's reference forms an attention query. The attended visual features and skill type determine one transition adjustment.
        (d) Residual adapters are initialized by behavior cloning and optimized with single-skill GRPO.
        Progress estimators use supervised learning, while the transition policy is trained with offline behavior cloning.
    }
    \label{fig:framework}
\end{figure*}

AeroSkill combines skill generation, execution, and transitions (Fig.~\ref{fig:framework}).
A frozen large language model (LLM) parses $L$ into an ordered skill sequence.
Skill-specific residual adapters refine actions from UAV-adapted OpenVLA~\cite{kim2025openvla} using state history, skill parameters, and predicted progress.
Progress also governs skill switching.
A visual transition policy uses the successor's reference information and skill type to predict one adjustment.
Sections~\ref{sec:parameterized_skills}--\ref{sec:training} detail these modules and their training.

\subsection{Parameterized Flight Skills}
\label{sec:parameterized_skills}

Flight operations in complex tasks share reusable control structures.
Based on motion reference relationships, trajectory constraints, and termination conditions, we define five skill types: turn, move, navigate, orbit, and land.
The skill type specifies the shared behavior, while explicit parameters describe its particular requirements.
The $i$-th skill instance is represented as
\begin{equation}
    z_i=(k_i,\theta_i),
    \label{eq:skill_instance}
\end{equation}
where $k_i$ denotes the skill type and $\theta_i$ contains the parameters associated with that type.
Figure~\ref{fig:flight_skills} shows the five flight skills and the task parameters used to specify each one.

The frozen LLM receives the task instruction $L$ and a prompt defining the available skill types, their parameter fields, and the required JSON format.
It parses the instruction into an ordered skill sequence.
Each JSON entry records a skill type and its extracted parameter text:
\begin{equation}
    \mathcal Z=\mathcal P_{\mathrm{LLM}}(L)=(z_1,z_2,\ldots,z_N),
    \label{eq:skill_sequence}
\end{equation}
where $\mathcal P_{\mathrm{LLM}}$ denotes instruction parsing and $N$ is the number of skill instances.
Parsing occurs once at the start of the task, after which the skills are executed in sequence.
The skill type and parameter text form the language input to OpenVLA.
The type selects the corresponding residual adapter and progress estimator.
A skill-specific mapping converts the extracted fields $\theta_i^{\mathrm{text}}$ into a vector, $\theta_i=E_{k_i}(\theta_i^{\mathrm{text}})$.
This vector conditions residual correction and progress prediction.
During a transition, the next skill's reference text supplies the query for visual attention.

\subsection{Progress-Aware Residual Control}
\label{sec:residual_control}

Environmental disturbances, perception errors, and deviations in control response can cause base actions to fall short of skill requirements.
To compensate for these deviations using execution feedback, we equip each of the five skill types with an MLP residual adapter on top of frozen OpenVLA.
For the current skill $z_i=(k_i,\theta_i)$, OpenVLA generates a base action $a_t^{\mathrm{VLA}}$ from the visual observation and skill text.
The corresponding adapter takes the base action, position history $H_t$, skill parameters, and predicted progress as inputs.
Its residual correction is combined with the base action to produce the final control action:
\begin{equation}
    a_t=a_t^{\mathrm{VLA}}
    +\alpha\,\pi_{\phi_{k_i}}^{\mathrm{res}}\!
    (a_t^{\mathrm{VLA}},H_t,\theta_i,\hat p_t).
    \label{eq:residual_action}
\end{equation}
Here, $\phi_{k_i}$ denotes the adapter parameters and $\alpha$ is a fixed scaling factor.
The network receives the vector representation of $\theta_i$.
The residual and base action share the same action space and coordinate frame.
The history $H_t$ contains the most recent $N_{\mathrm{hist}}$ position states, including the current state.
We reset this history at the start of each new skill.

Remaining skill requirements change during execution.
An independent MLP for each skill type predicts completion progress from OpenVLA visual encoder features $v_t$, position history, and task parameters:
\begin{equation}
    \hat p_t=g_{\psi_{k_i}}(v_t,H_t,\theta_i),
    \label{eq:progress_prediction}
\end{equation}
where $\psi_{k_i}$ denotes the parameters of the corresponding estimator.
Target parameters distinguish instances of the same skill type with different distance, angle, or other requirements.
For skills with spatial paths, supervision is derived by projecting the current position onto the ground-truth (GT) path from the dataset.
Let $\Gamma_i(\ell)$ denote an arc-length-parameterized path, where $\ell\in[0,L_i]$ and $L_i>0$ is the total path length.
Using normalized arc length $u\in[0,1]$, we define the label at $\mathbf s_t$ as
\begin{equation}
    p_t^{\mathrm{GT}}
    =\underset{u\in[0,1]}{\arg\min}\,
    \left\|\mathbf s_t-\Gamma_i(L_i u)\right\|_2^2.
    \label{eq:gt_progress}
\end{equation}
For Orbit, we initialize $p_0^{\mathrm{GT}}=0$ and enforce nondecreasing progress, $p_t^{\mathrm{GT}}\ge p_{t-1}^{\mathrm{GT}}$.
The temporal order along the GT trajectory distinguishes repeated positions.
For Turn, progress labels use the ratio of accumulated rotation to the GT target angle.
Predicted progress conditions residual control and triggers switching when $\hat p_t>\tau_p$.
The default threshold is $\tau_p=0.95$. We determine whether a task succeeds using independent evaluation criteria.

\subsection{Visually Grounded Skill Handoffs}
\label{sec:transition_policy}

A skill may end in a state unsuitable for its successor.
For example, an orbit can end at the required radius while leaving the UAV facing away from the landmark for the next maneuver.
The handoff depends on the flight state reached at skill completion and on what the successor needs from that state.
The successor's reference guides visual attention to scene features used to compute the adjustment.

For the successor skill $z_{i+1}=(k_{i+1},\theta_{i+1})$, we extract its reference text $r_{i+1}$ from the JSON entry.
The reference encoder forms the attention query as $Q_{i+1}=f_{\mathrm{ref}}(r_{i+1})$.
Current scene features from the frozen OpenVLA visual encoder provide cross-attention keys and values:
\begin{equation}
    c_t^{i+1}=\operatorname{softmax}\!\left(
        \frac{Q_{i+1}K_t^\top}{\sqrt d}
    \right)V_t.
    \label{eq:transition_attention}
\end{equation}
Here, $Q_{i+1}$ is the reference query, $K_t$ and $V_t$ are the visual keys and values, and $d$ is the query and key dimension.
The policy combines the attended representation with the successor type embedding:
\begin{equation}
    a_t^{\mathrm{trans}}=\pi_{\omega}^{\mathrm{trans}}\!\left(
        c_t^{i+1},e_{k_{i+1}}
    \right),
    \label{eq:transition_action}
\end{equation}
where $\omega$ denotes the policy parameters and $e_{k_{i+1}}$ is the skill type embedding of the successor skill.
After one adjustment, control passes to the successor.
Turn and Move bypass this step because their heading or displacement commands specify the motion without a reference object.

The transition policy is trained through offline behavior cloning (BC).
We compute the deviation between each executed endpoint and its GT endpoint in body coordinates.
To perturb a successor's GT start, we apply the same offset components along the body axes at that start.
For example, a forward offset follows the successor's starting heading.
The inputs are the observations at the perturbed starts, the successor's reference text, and its skill type.
The target is a one-step action toward the reference start in the body-frame action space of Eq.~\eqref{eq:action_space}.
The objective minimizes the mean squared error (MSE) between predicted and target actions.
The visual encoder remains frozen throughout transition policy training, and only the trainable parameters of the transition module are optimized.
GT starting and terminal states are used to construct offline supervision.
At deployment, the policy generates actions from current observations, the successor's reference text, and its skill type.

\subsection{Training Objectives and Optimization}
\label{sec:training}

We initialize each residual adapter by BC. We then refine it on individual skills using online group relative policy optimization (GRPO)~\cite{shao2024deepseekmath}.
During BC, we minimize the MSE between the combined action in Eq.~\eqref{eq:residual_action} and the demonstration action $a_t^{\mathrm{GT}}$.
Equivalently, Fig.~\ref{fig:framework}(d) shows the target $a_t^{\mathrm{GT}}-a_t^{\mathrm{VLA}}$ for the scaled correction $\alpha\pi_{\phi_{k_i}}^{\mathrm{res}}$.
OpenVLA remains frozen.
For skills involving position changes, the reward combines GT path advancement and deviation penalties, with
$\Delta \mathbf{s}_t = \mathbf{s}_{t+1} - \mathbf{s}_t$:
\begin{equation}
    r_t =
    w_f \frac{\Delta \mathbf{s}_t^\top \mathbf{u}_t}{\ell_{\mathrm{ref}}}
    - w_d \left(1 - e^{-d_{t+1}/\sigma_d}\right).
\end{equation}
Here, $\mathbf u_t$ is the directed unit tangent at the projection of the current position onto the GT path.
The distance $d_{t+1}$ is the shortest distance from the next position to that path.
The parameters $\ell_{\mathrm{ref}}$ and $\sigma_d$ control displacement normalization and distance decay, respectively, while $w_f,w_d>0$ are reward weights.
The forward term rewards actual displacement along the reference direction, and the deviation term penalizes departures from the path.
For Turn, an angular error reward replaces the translational forward term.
It compares the actual accumulated rotation angle with the GT target angle, assigning higher rewards to smaller errors.

During training, continuous residual actions are sampled from a learnable Gaussian policy $q_{\phi_k}^{\mathrm{res}}(\Delta a_t\mid c_t)$, conditioned on
\begin{equation}
    c_t=(a_t^{\mathrm{VLA}},H_t,\theta_i,\hat p_t).
    \label{eq:residual_condition}
\end{equation}
At deployment, the policy mean is used as the deterministic residual.
For the same skill instance and initial state, the old policy samples $G$ trajectories.
The trajectory return and group-relative advantage are defined as
\begin{equation}
    R_j=\sum_{t=0}^{T_j-1}\gamma^t r_{j,t},
    \qquad
    A_j=\frac{R_j-\bar R}{\sigma_R+\delta},
    \label{eq:group_advantage}
\end{equation}
where $\bar R$ and $\sigma_R$ are the mean and standard deviation of returns within the group.
The constant $\delta>0$ ensures numerical stability.
Let $\rho_{j,t}$ denote the probability density ratio between the new and old residual policies for the same sampled action and conditioning inputs.
Define the clipped ratio as $\bar\rho_{j,t}=\operatorname{clip}(\rho_{j,t},1-\epsilon,1+\epsilon)$.
The clipped surrogate is $\mathcal S_{j,t}=\min\{\rho_{j,t}A_j,\bar\rho_{j,t}A_j\}$.
We maximize the averaged objective
\begin{equation}
    \mathcal J_{\mathrm{GRPO}}(\phi_k)
    =\mathbb E\!\left[
        \frac{1}{G}\sum_{j=1}^{G}
        \frac{1}{T_j}\sum_{t=0}^{T_j-1}\mathcal S_{j,t}
    \right].
    \label{eq:grpo_objective}
\end{equation}
Here, $T_j$ is the trajectory length, $\gamma$ is the discount factor, and $\epsilon$ is the clipping threshold.
GRPO updates only the residual adapter of the current skill.
Progress estimators use MSE supervision on demonstrations, then on interaction states with GT progress labels.
Their updates are independent of GRPO.
Residual control uses predicted progress in both training and deployment.
Together, these modules support long-horizon execution by training skills with online interaction and learning the transition policy offline.

\section{Experiments}
\label{sec:experiments}

We evaluate long-horizon completion, single-skill precision, module contributions, and robustness to skill-entry and mid-skill perturbations.
We also test real-world deployment without deployment-specific fine-tuning.

\subsection{Experimental Setup}
\label{sec:exp_setup}

\paragraph{Benchmarks and Metrics.}
We evaluate 200 long-horizon tasks comprising 697 subtasks from the Long-horizon Flow split of FLIGHT~\cite{zheng2026flight}, and ten instruction categories across five skill families on UAV-Flow~\cite{wang2025uavflow}.
For both benchmarks, we manually verify instruction compliance throughout execution, checking both the flight process and outcome.
Examples include the specified passing side, completion of the prescribed orbit, and the relative landing position.

On FLIGHT, we manually assess each subtask against its instruction.
Following the benchmark protocol, a task succeeds only if all required subtasks succeed.
Task success rate (SR) is the fraction of successful tasks, while stage success rate (SSR) is the fraction of successful subtasks across all tasks, counting unreached and timed-out subtasks as failures.
On UAV-Flow, success requires both passing the manual assessment of instruction compliance and achieving a final position error below 3\,m and a final yaw error below $10^\circ$.
Normalized dynamic time warping (nDTW)~\cite{ilharco2019ndtw} measures agreement with reference trajectories on both benchmarks.
We report UAV-Flow SR and nDTW per category, with overall scores weighted by trial count.

\paragraph{Implementation.}
We use Qwen3.8-Max to parse instructions into ordered sequences of parameterized skills.
Training uses two NVIDIA RTX 6000 Ada GPUs with 48\,GB memory each.
The controller runs at 5\,Hz with an eight-step position history and a residual scale of $\alpha=0.1$.
Residual adapters are initialized with BC and refined through single-skill GRPO, while the pretrained UAV-VLA backbone remains frozen throughout both stages.
Progress estimators use independent supervision, while the transition policy is trained with offline BC.
Training and test splits share scenes, but FLIGHT test trajectories and instructions are held out from BC, GRPO, and transition training.

\subsection{Long-Horizon Task Execution}
\label{sec:exp_long_horizon}

\begin{table}[t]
\centering\small
\setlength{\fboxsep}{0pt}
\caption{\textbf{Long-horizon results and ablations on FLIGHT.} Published baseline scores are from~\cite{zheng2026flight}. The \textbf{best} and second-best scores in each column are highlighted in \protect\colorbox{red!15}{\rule{0pt}{0.65em}\hspace{0.65em}} and \protect\colorbox{orange!25}{\rule{0pt}{0.65em}\hspace{0.65em}}, respectively.}
\label{tab:exp_flight}

\setlength{\fboxsep}{1.2pt}
\setlength{\tabcolsep}{1.8pt}
\renewcommand{\arraystretch}{0.98}

\begin{tabular*}{\columnwidth}{@{\extracolsep{\fill}}lccc@{}}
\toprule
\textbf{Method}
& \textbf{SR (\%)}$\uparrow$
& \textbf{SSR (\%)}$\uparrow$
& \textbf{nDTW} ($\times 100$)$\uparrow$ \\
\midrule
OpenVLA-UAV
& 10.50 & 30.87 & 18.02 \\
Memory VLA
& 18.50 & 38.69 & 16.58 \\
FLIGHT VLA
& 59.00 & \colorbox{orange!25}{78.97} & 45.78 \\
\midrule
AeroSkill
& \colorbox{red!15}{\textbf{70.50}}
& \colorbox{red!15}{\textbf{84.22}}
& \colorbox{red!15}{\textbf{53.64}} \\
\hspace{0.4em}w/o residual correction
& 61.00 & 77.47 & 41.03 \\
\hspace{0.4em}w/o transition policy
& \colorbox{orange!25}{64.50} & 73.03 & 46.42 \\
\hspace{0.4em}w/o residual \& transition
& 49.50 & 58.68 & 38.35 \\
\hspace{0.4em}w/o progress input
& 62.00 & 77.91 & \colorbox{orange!25}{47.23} \\
\hspace{0.4em}w/o progress switching
& 33.50 & 51.36 & 29.18 \\
\hspace{0.4em}w/o progress module
& 27.50 & 46.63 & 21.12 \\
\bottomrule
\end{tabular*}
\end{table}

We test AeroSkill on instructions that require several skills in a single flight.
The UAV executes the skills in order.
After each transition, the next skill starts from the resulting flight state.
Table~\ref{tab:exp_flight} compares AeroSkill with published FLIGHT baselines using the same test tasks and evaluation budget.
AeroSkill achieves 70.50\% task SR, exceeding FLIGHT VLA by 11.50 percentage points.
SSR rises from 78.97\% to 84.22\%, while nDTW improves from 45.78 to 53.64.
The larger gain in task SR than SSR indicates that more flights satisfy every requirement of the instruction.
nDTW also indicates closer trajectory agreement.

\subsection{Single-Skill Execution Precision}
\label{sec:exp_single_skill}

UAV-Flow tests each skill in isolation under its specified motion requirements, separating execution quality from handoff effects.
We compare the frozen OpenVLA-UAV backbone with residual adapters trained by BC alone or BC followed by GRPO, using the same backbone.
Overall SR rises from 60.07\% to 71.06\% after BC and reaches 79.12\% after GRPO (Table~\ref{tab:exp_uavflow}).
The gains vary by category: BC improves Land from 44.44\% to 61.11\% but reduces Pass from 37.50\% to 32.50\%; GRPO raises Pass to 55.00\%.

The full skill executor also improves overall nDTW from 34.42 to 56.64.
On Asc./Desc., all three variants achieve 100\% SR, yet nDTW rises from 72.94 to 96.63 over the backbone.
Thus, successful execution can still differ in trajectory fidelity.
Pass improves in both SR and nDTW, showing gains in instruction compliance and reference-path agreement.
These results further support using the refined skills as reusable building blocks for longer tasks.

\begin{table*}[t]
    \centering
    \caption{\textbf{Single-skill execution performance on UAV-Flow.} We report SR and nDTW. Overall scores are weighted by trial count. The result highlighting scheme follows Table~\ref{tab:exp_flight} and is applied only to the Overall column.}
    \label{tab:exp_uavflow}
    \small
    \setlength{\tabcolsep}{2pt}
    \setlength{\fboxsep}{1.2pt}
    \renewcommand{\arraystretch}{1.02}
    \begin{tabular*}{\textwidth}{@{\extracolsep{\fill}}l*{10}{c}c@{}}
        \toprule
        \textbf{Skill}
        & \multicolumn{2}{c}{Turn}
        & \multicolumn{2}{c}{Move}
        & \multicolumn{4}{c}{Navigate}
        & Land & Orbit & \multicolumn{1}{c@{}}{} \\
        \cmidrule(lr){2-3}
        \cmidrule(lr){4-5}
        \cmidrule(lr){6-9}
        \cmidrule(lr){10-10}
        \cmidrule(lr){11-11}
        \textbf{UAV-Flow task type}
        & Rotate & Turn & Shift & Asc./Desc.
        & Retreat & Move & Pass & Approach
        & Land & Surround & \textbf{Overall} \\
        \midrule
        \multicolumn{12}{@{}l@{}}{\textbf{SR (\%)}$\uparrow$} \\
        Frozen OpenVLA-UAV
        & 13.33 & \textbf{100.00} & \textbf{100.00} & \textbf{100.00}
        & 83.33 & \textbf{100.00} & 37.50 & 33.33
        & 44.44 & 8.33 & 60.07 \\
        Residual (BC only)
        & \textbf{100.00} & 73.33 & 97.96 & \textbf{100.00}
        & \textbf{91.67} & \textbf{100.00} & 32.50 & 40.48
        & 61.11 & \textbf{100.00} & \colorbox{orange!25}{71.06} \\
        Residual (BC+GRPO)
        & \textbf{100.00} & \textbf{100.00} & 97.96 & \textbf{100.00}
        & \textbf{91.67} & \textbf{100.00} & \textbf{55.00} & \textbf{54.76}
        & \textbf{66.67} & \textbf{100.00}
        & \colorbox{red!15}{\textbf{79.12}} \\
        \midrule
        \multicolumn{12}{@{}l@{}}{\textbf{nDTW} ($\times 100$)$\uparrow$} \\
        Frozen OpenVLA-UAV
        & 33.40 & 21.87 & 71.22 & 72.94
        & 29.67 & 15.60 & 21.27 & 15.74
        & 12.71 & 75.30 & 34.42 \\
        Residual (BC only)
        & \textbf{90.23} & 53.01 & 65.58 & 68.02
        & 51.69 & 21.50 & 32.40 & 19.63
        & 22.59 & 66.78 & \colorbox{orange!25}{43.00} \\
        Residual (BC+GRPO)
        & 89.45 & \textbf{68.65} & \textbf{95.10} & \textbf{96.63}
        & \textbf{60.24} & \textbf{27.61} & \textbf{44.96} & \textbf{31.27}
        & \textbf{24.25} & \textbf{86.45}
        & \colorbox{red!15}{\textbf{56.64}} \\
        \bottomrule
    \end{tabular*}
\end{table*}

\subsection{Ablations and Robustness}
\label{sec:exp_analysis}

\subsubsection{Module Ablations on FLIGHT}

We first evaluate a minimal backbone-chaining baseline to establish the performance of the shared composition framework.
We remove both residual correction and transition adjustment while retaining the same parser, skill inputs, and progress-based switching.
The remaining learned weights are fixed without compensatory training.
This baseline achieves 49.50\% task SR, compared with 10.50\% for the published OpenVLA-UAV baseline.
This supports the effectiveness of the basic composition framework without either module.
Restoring either module improves all three metrics.
The full model reaches 70.50\% SR, which is 21.00 percentage points above backbone chaining (Table~\ref{tab:exp_flight}).

\begin{figure}[t]
\centering
\includegraphics[width=\linewidth]{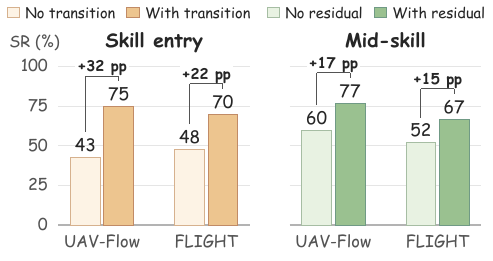}
\caption{\textbf{SR under state perturbations.} UAV-Flow reports SR for individual skills. FLIGHT reports task SR for complete flights. Skill-entry perturbations test the transition policy. Mid-skill perturbations test residual correction.}
\label{fig:exp_robustness}
\end{figure}

\paragraph{Residual Correction.}
Residual correction addresses motion deviations during skill execution.
We disable it while retaining progress-based switching and transition adjustment.
Task SR and SSR fall by 9.50 and 6.75 percentage points, respectively, while nDTW drops from 53.64 to 41.03 (Table~\ref{tab:exp_flight}).
The lower task completion and trajectory agreement indicate that the frozen backbone benefits from action correction when its skills are executed consecutively.
The lower SSR indicates that residual correction supports subtask completion across the evaluated tasks.

\paragraph{Transition Policy.}
Completing one skill does not necessarily provide a suitable starting state for its successor.
We compare transition adjustment with direct switching, keeping the skill executors and switching rule unchanged.
Task SR rises from 64.50\% to 70.50\%, SSR by 11.19 percentage points, and nDTW by 7.22 points (Table~\ref{tab:exp_flight}).
These improvements support adjusting handoff states to help subsequent skills succeed, without changing the skill executors.
Transition adjustment lifts SSR above the FLIGHT VLA baseline. Direct switching leaves SSR below the reported baseline.

\paragraph{Progress Module.}
Progress guides residual correction and decides when to switch skills.
First, we remove progress from the residual input and retrain the adapters, keeping progress-based switching unchanged.
Task SR drops from 70.50\% to 62.00\%.
Next, we retain progress-conditioned residual control and replace progress-based switching with a rule adapted from UAV-Flow~\cite{wang2025uavflow}.
This rule ends a skill after ten consecutive steps with displacement below 10\,cm and yaw change below $3^\circ$ at each step, or at the step limit.
With learned weights fixed, task SR falls to 33.50\%.
Removing progress from both correction and switching reduces SR to 27.50\% and SSR to 46.63\% (Table~\ref{tab:exp_flight}).
Both individual ablations also lower SSR and nDTW, supporting the use of progress in residual control and skill switching.

\begin{figure*}[t]
    \centering
    \includegraphics[width=\textwidth]{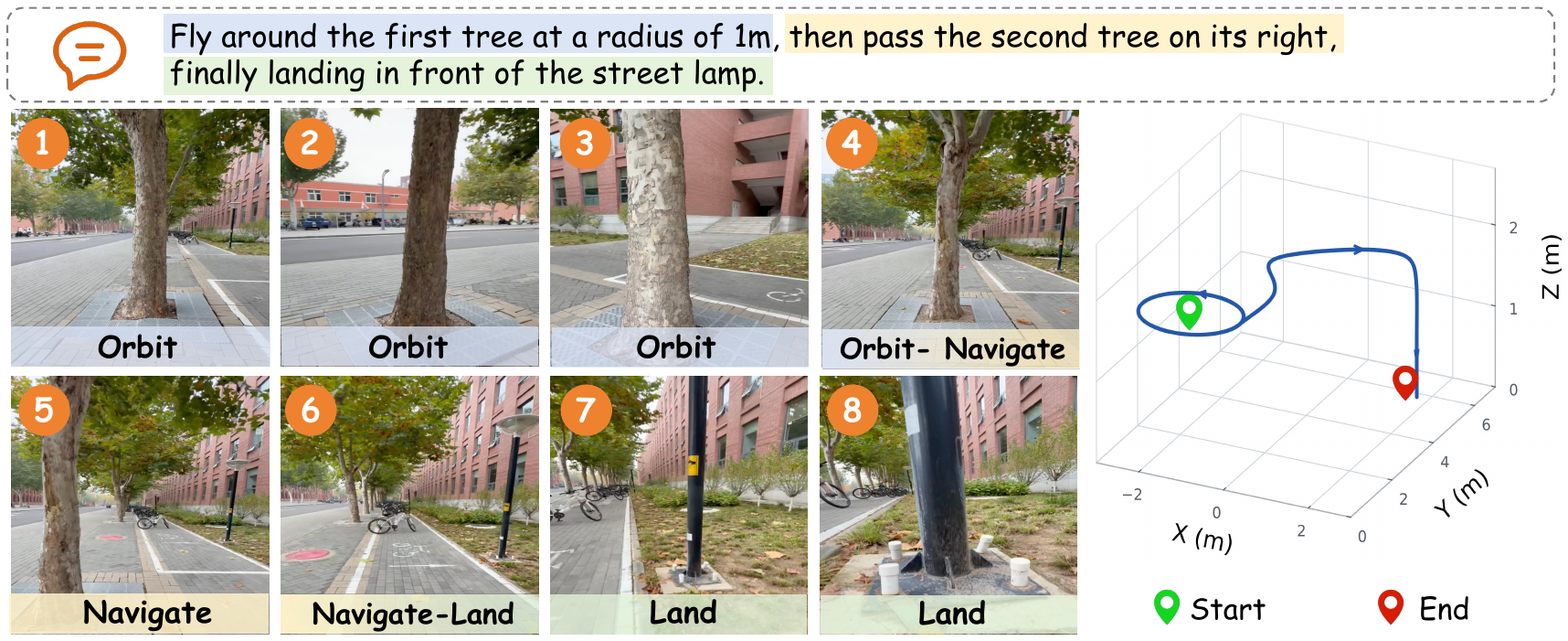}
    \caption{\textbf{Real-world execution with AeroSkill.} The UAV orbits the first tree (Orbit). It then passes the second tree on the right (Navigate) and lands in front of the street lamp (Land). The numbered keyframes show the sequence. The 3D path is on the right.}
    \label{fig:exp_realworld}
\end{figure*}

\subsubsection{Robustness to State Perturbations}
\label{sec:exp_recovery}

\paragraph{Handoff Robustness.}
Skill-entry perturbations test whether one adjustment can compensate for mismatches between actual and reference starting states.
On UAV-Flow, we perturb position and yaw at skill entry, limiting translation to 0.8\,m.
Both variants use the same skill executor; one adds a single transition adjustment.
The adjustment raises SR from 43\% to 75\% and nDTW from 37.15 to 54.11.
A separate FLIGHT full-task test with matched skill-entry perturbations raises task SR from 48\% to 70\% (Figure~\ref{fig:exp_robustness}).
These results support using a single adjustment to improve execution from perturbed skill-entry states.

\paragraph{Mid-Skill Recovery.}
Perturbations introduced after skill entry test feedback correction during ongoing execution.
We apply random state perturbations within skills and compare AeroSkill with and without residual correction.
Residual correction raises UAV-Flow SR from 60\% to 77\% and FLIGHT task SR from 52\% to 67\% (Figure~\ref{fig:exp_robustness}).
The gains in both settings support residual correction for handling disturbances during individual skills and their composition.
On UAV-Flow the gains reflect individual skill execution, while on FLIGHT they reflect success over complete flights.

\par\medskip
\noindent\begin{minipage}{\columnwidth}
\centering
\includegraphics[width=\linewidth]{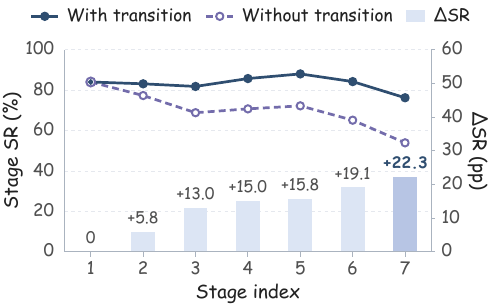}
\captionof{figure}{\textbf{Subtask SR by position on FLIGHT.} The lines show mean SR across tasks with and without the transition policy (left axis). The bars show the SR difference with minus without the policy (right axis, percentage points).}
\label{fig:exp_stagewise}
\end{minipage}\par

\subsubsection{Transition Reliability Across Stages}
\label{sec:exp_transitions}
We further examine whether the transition benefit persists across successive skills.
At position $k$, we consider tasks containing at least $k$ subtasks and compute each task's $k$-th subtask SR using only trials that reach it.
We average these rates across tasks, separately for each variant, excluding tasks with no eligible trials.
Figure~\ref{fig:exp_stagewise} shows these conditional means and their differences.
Both variants achieve 84.0\% at the first subtask, before any transition.
With transition adjustment, mean SR is higher at all later positions; the gap grows from 5.8 percentage points at the second to 22.3 at the seventh (76.2\% versus 53.9\%).
These conditional results indicate benefits at later stages.

\subsection{Real-World Evaluation}
\label{sec:exp_realworld}

We deploy AeroSkill on a FanciSwarm Pro UAV. A server with an NVIDIA RTX 6000 Ada GPU runs inference and communicates with the UAV over a local Wi-Fi network. 
Neither system receives deployment-specific fine-tuning.

Each system flies once on the same 100 long-horizon instructions.
AeroSkill executes parsed skills, while OpenVLA-UAV receives the full instruction directly.
In this complete-system comparison, AeroSkill completes 51/100 tasks versus 7/100 and achieves 62\% SSR versus 24\% (Table~\ref{tab:exp_realworld}).
These results support the complete framework in real flights; the individual modules are examined separately in the simulation ablations.
Figure~\ref{fig:exp_realworld} shows consecutive skills using different landmarks. At each handoff, the successor starts from the UAV's current flight state.

\begin{table}[!htb]
\centering\small
\caption{\textbf{Real-world results on the same 100 instructions.}}
\label{tab:exp_realworld}
\setlength{\tabcolsep}{3pt}
\begin{tabular*}{\columnwidth}{@{\extracolsep{\fill}}lcc@{}}
\toprule
\textbf{Method} & \textbf{Task SR(\%)}$\uparrow$ & \textbf{SSR (\%)}$\uparrow$ \\
\midrule
OpenVLA-UAV & 7/100 (7\%) & 24.00\\
AeroSkill & \textbf{51/100 (51\%)} & \textbf{62.00}\\
\bottomrule
\end{tabular*}
\end{table}

\section{Conclusion}
\label{sec:conclusion}

AeroSkill composes parameterized flight skills with a frozen UAV-VLA model.
Progress guides residual correction and skill switching, while visual transitions prepare suitable starting states.
Single-skill interaction and offline transition learning avoid full-task online training.
Tests in simulation and real flights show improved skill execution and higher long-horizon task success rates.

\par\pagebreak
\clearpage

{\small
\bibliographystyle{ieeenat_fullname}
\bibliography{main}
}

\flushcolsend
\clearpage
\raggedcolsend
\appendix
\setcounter{figure}{0}
\setcounter{table}{0}
\setcounter{equation}{0}
\renewcommand{\thefigure}{S\arabic{figure}}
\renewcommand{\thetable}{S\arabic{table}}
\renewcommand{\theequation}{S\arabic{equation}}
\renewcommand{\theHfigure}{appendix.\arabic{figure}}
\renewcommand{\theHtable}{appendix.\arabic{table}}
\renewcommand{\theHequation}{appendix.\arabic{equation}}

\twocolumn[{%
  \centering\Large\bfseries Appendix\par\vspace{12pt}
}]
\lstset{basicstyle=\ttfamily\footnotesize,columns=fullflexible,keepspaces=true,
  breaklines=true,frame=single,rulecolor=\color{black!25}}

\small
\section{Instruction Parsing and Skill Execution}
\subsection{Parser prompt and output schema}

The frozen Qwen3.8-Max parser is called once before flight. It receives the system prompt below and the natural-language task instruction. The prompt is lightly edited for readability.

\begin{lstlisting}
Parse the UAV instruction once before flight.
Return only the following JSON object:
{"skills":[{"index":0,"type":"<skill type>",
            "parameters":{...}}]}
Use exactly these five lowercase types:
turn (change heading in place):
  direction (left | right), angle
move (fly in the specified direction):
  direction (forward | backward | left |
             right | up | down), distance
navigate (pass or approach an object):
  reference, position (left | right | front |
                       behind)
orbit (circle an object): reference, radius
land (land relative to an object):
  reference, position (left | right | front |
                       behind)
Emit one entry per motion stage in the original
order. Use consecutive indices starting at 0.
Attach perception phrases to the associated
motion. With no motion, return {"skills":[]}.
Keep object descriptions and modifiers intact.
Copy quantities with their original units, such
as "90 degrees", "5 m", and "3 m". Omit an
unspecified optional field or use null.
Do not compute values or supply defaults.
Check types, indices and permitted keys. Require
a reference for navigate, orbit, and land.
Return JSON without Markdown or commentary.
\end{lstlisting}

\noindent\begin{minipage}{\columnwidth}
\subsection{Example and execution interface}

\textit{Instruction.} Fly around the tree at a radius of 1 m, then pass the car on its right and land to the right of the streetlamp.

\begin{lstlisting}
{"skills":[
  {"index":0,"type":"orbit","parameters":
    {"reference":"tree","radius":"1 m"}},
  {"index":1,"type":"navigate","parameters":
    {"reference":"car","position":"right"}},
  {"index":2,"type":"land","parameters":
    {"reference":"streetlamp","position":"right"}}
]}
\end{lstlisting}
\end{minipage}

The executor validates the JSON against the schema and advances a pointer through the fixed skill list. The type selects a residual adapter and a progress estimator. The type and parameter text form the backbone instruction, for example \textit{Orbit the tree at a radius of 1 m}. A skill-specific mapping converts the parameter text to the numerical representation consumed by the learned modules. At handoff, the transition module uses the successor's reference text to form its attention query over the current visual features.

\section{Training and Evaluation Details}
\subsection{Residual behavior cloning and GRPO}

\paragraph{Behavior cloning.}
The demonstration action supervises the sum of the frozen OpenVLA~\cite{kim2025openvla} action and the scaled residual. The unscaled residual target is $(a_t^{\mathrm{GT}}-a_t^{\mathrm{VLA}})/\alpha$, where $\alpha=0.1$. The resulting adapter provides the starting policy for single-skill refinement with group relative policy optimization (GRPO)~\cite{shao2024deepseekmath}.

\paragraph{Single-skill rollout collection.}
Each iteration samples 54 skill instances. For each instance, the current policy generates four trajectories from the same initial state, giving 216 trajectories per iteration. At step $t$, the residual policy receives
$c_t=(a_t^{\mathrm{VLA}},H_t,\theta_i,\hat p_t)$,
where $a_t^{\mathrm{VLA}}$ is the frozen backbone action, $H_t$ is the position history, $\theta_i$ contains the skill parameters, and $\hat p_t$ is the estimated progress. The Gaussian policy draws a correction $\Delta a_t$, and the environment receives $a_t=a_t^{\mathrm{VLA}}+\alpha\Delta a_t$. At each step of the rollout, we record the conditioning inputs, sampled correction, its log probability under the old policy, and the reward.

\paragraph{Per-step rewards.}
For translating skills, let $\Delta s_t=s_{t+1}-s_t$ be the executed displacement, $u_t$ the directed unit tangent at the projection of the current position onto the reference path, and $d_{t+1}$ the distance from the next position to that path. The reward is
\begin{equation}
r_t = w_f\frac{\Delta s_t^\top u_t}{\ell_{\mathrm{ref}}}
      -w_d\bigl(1-e^{-d_{t+1}/\sigma_d}\bigr).
\end{equation}
The first term uses executed motion projected along the reference direction. The second penalizes deviation after the action. The scales $\ell_{\mathrm{ref}}$ and $\sigma_d$ normalize displacement and deviation, respectively. For Turn, an angular reward replaces the forward term. It compares the accumulated rotation with the target angle and gives the policy a higher reward when the angular error is smaller.

\paragraph{Within-group advantages.}
For rollout $j$ of length $T_j$, we compute its discounted return $R_j$ and standardize it against the other rollouts of that skill instance:
\begin{align}
R_j &= \sum_{t=0}^{T_j-1}\gamma^t r_{j,t}, &
A_j &= \frac{R_j-\bar R}{\sigma_R+\delta},
\end{align}
where $\bar R$ and $\sigma_R$ are the group mean and standard deviation. Each sampled correction inherits the advantage of its trajectory.

\paragraph{Adapter update.}
For each recorded sample, the updated policy recomputes the log probability of the same correction under the same conditioning inputs. The ratio $\rho_{j,t}$ is the exponential of the new minus old log probability. With $\bar\rho_{j,t}=\operatorname{clip}(\rho_{j,t},1-\epsilon,1+\epsilon)$, the per-step objective is
\begin{equation}
S_{j,t}=\min\!\left(\rho_{j,t} A_j,\bar\rho_{j,t} A_j\right).
\end{equation}
We average $S_{j,t}$ within each trajectory before averaging over the four trajectories. This gives each trajectory equal weight even when rollout lengths differ. The update modifies only the current skill adapter. After optimization, the updated policy collects a new rollout batch for the next iteration. Table~\ref{tab:grpo-config} summarizes the schedule. At deployment, the controller uses the mean of the learned correction distribution as its deterministic residual action.

\begin{table}[ht]
\centering
\caption{GRPO rollout schedule and optimizer settings.}\vspace{2pt}
\label{tab:grpo-config}
\small
\begin{tabularx}{\columnwidth}{@{}Xl@{}}
\toprule
Setting & Value \\
\midrule
Skill instances sampled per iteration & 54 \\
Rollouts per instance & 4 \\
Trajectories per iteration / per adapter & 216 / 648 \\
Iterations per adapter & 3 \\
Optimization epochs by iteration & 2, 2, 1 \\
Optimizer & Adam \\
Learning rate & $10^{-6}$ \\
Minibatch size & 512 \\
Importance-ratio clipping $\epsilon$ & 0.2 \\
Maximum gradient norm & 0.5 \\
\bottomrule
\end{tabularx}
\end{table}

\subsection{Independent progress supervision}

The progress estimator is a separate MLP trained with mean squared error on demonstrations and labeled interaction states. Spatial labels use normalized arc length at the reference-path projection. Orbit labels follow trajectory order and are nondecreasing. Turn labels use the ratio of accumulated rotation to the target angle. The predicted progress enters the residual adapter and triggers a handoff once the prediction exceeds the switching threshold.

\begin{table}[ht]
\centering
\caption{Shared controller settings.}
\small
\begin{tabularx}{\columnwidth}{@{}lX@{}}
\toprule
Setting & Specification \\
\midrule
Action & Body-frame position and yaw increments \\
Control frequency & $5$ Hz \\
Position history & $8$ states including the current state \\
Residual scale & $\alpha=0.1$ \\
Switch condition & Predicted progress above $0.95$ by default \\
\bottomrule
\end{tabularx}
\end{table}

\subsection{Transition policy supervision}

For offline behavior cloning, the offset between an executed endpoint and its reference endpoint is expressed in body coordinates. Its components are transferred to the body axes at a successor's reference start, so a forward offset follows the successor's starting heading. The input consists of the observation at the perturbed start, the successor's reference text, and its skill type. The target is a one-step action toward the reference start, expressed in the body-frame action space. The visual encoder remains frozen during this training. At deployment, the adjustment is applied once before an eligible successor begins. Turn and Move bypass the adjustment. The position history is reset at the beginning of the next skill.

\subsection{Perturbation protocols}

\textbf{Skill entry.} UAV-Flow applies a pose perturbation at the start of a single-skill trial, with translation limited to $0.8$ m as reported in the main paper. FLIGHT applies perturbations at skill starts in the sequence. Both conditions use the same skill executor. In the transition-enabled condition, the controller applies a single adjustment before executing a skill that specifies a reference object.

\textbf{Mid-skill.} A pose perturbation is injected during execution. The perturbed state enters the existing position history, and execution continues with fixed skill parameters. The two conditions differ in whether residual correction is active. We inject the perturbation at step 30 on UAV-Flow. On FLIGHT, the injection step is sampled from the interval $[25,35]$ during skill execution.

\textbf{Sampling.} The perturbation protocol draws Gaussian offsets in world-frame position and yaw, with per-axis position standard deviation $0.5$ m and yaw standard deviation $15^\circ$. Yaw offsets are clipped to $\pm30^\circ$. Paired conditions share task-level seeds. With base seed $b$ and zero-based skill index $j$, the documented seeds for the two injection modes are $b+j$ and $b+j+1000$, respectively.

\subsection{Evaluation sets}
UAV-Flow~\cite{wang2025uavflow} contains 273 test task files. The single-skill evaluation covers ten instruction categories. The FLIGHT~\cite{zheng2026flight} long-horizon split (\texttt{eval\_gt\_lhf}) contains 200 tasks and 697 subtasks. Its reference paths have mean and median lengths of $33.84$ m and $20.19$ m, with a maximum of $204.88$ m. The seven evaluation scenes and their task counts are AsianMedievalCity (20), Greek\_Island (17), MiddleEast (26), ModularNeighborhood (39), SuburbNeighborhood\_Day (25), Tokyo (28), and Venice (45).

\subsection{Evaluation metrics}
Let $s_i$ indicate task success, $m_i$ be the number of required subtasks, and $u_{ij}$ indicate success on subtask $j$ of task $i$. The task and stage success rates are
\begin{align}
\mathrm{SR} &= \frac{100}{N}\sum_{i=1}^{N}s_i,\\
\mathrm{SSR} &= 100\frac{\sum_i\sum_{j=1}^{m_i}u_{ij}}{\sum_i m_i}.
\end{align}
On FLIGHT, a task succeeds if every required subtask succeeds. Unreached and timed-out subtasks count as failures in SSR. For the stage-wise plot, each task's success rate is computed from trials that reach that stage. We average these rates over the eligible tasks, using the set of tasks that reach the stage for each variant. 

On UAV-Flow, success requires instruction compliance, final position error below $3$ m, and final yaw error below $10^\circ$. Overall scores pool the ten categories, weighting each by its trial count.

For a reference path $R_i$ of $L_i$ waypoints and an executed path $P_i$, DTW~\cite{ilharco2019ndtw} minimizes the sum of waypoint distances over monotone alignments $\pi$:
\begin{equation}
\mathrm{DTW}(R_i,P_i)=\min_{\pi}\sum_{(j,k)\in\pi}d(r_{ij},p_{ik}).
\end{equation}
With benchmark distance scale $d_{\mathrm{th}}$ and included trials $\mathcal I$,
\begin{equation}
\mathrm{nDTW}=\frac{100}{|\mathcal I|}\sum_{i\in\mathcal I}
\exp\!\left(-\frac{\mathrm{DTW}(R_i,P_i)}{L_i d_{\mathrm{th}}}\right).
\end{equation}

\flushcolsend
\onecolumn
\raggedcolsend
\normalsize

\newcommand{\casefigure}[3]{%
  \par\noindent\begin{minipage}{\textwidth}
  \centering\includegraphics[width=#1\textwidth]{figures/appendix/#2.pdf}
  \captionof{figure}{#3}
  \end{minipage}\par\medskip
}

\section{Simulation Examples}
The numbered keyframes follow execution order. Labels between skills mark handoffs within each sequence. Section~\ref{sec:qualitative-perturbation} shows the effects of perturbations injected at skill entry and during execution.

\casefigure{0.96}{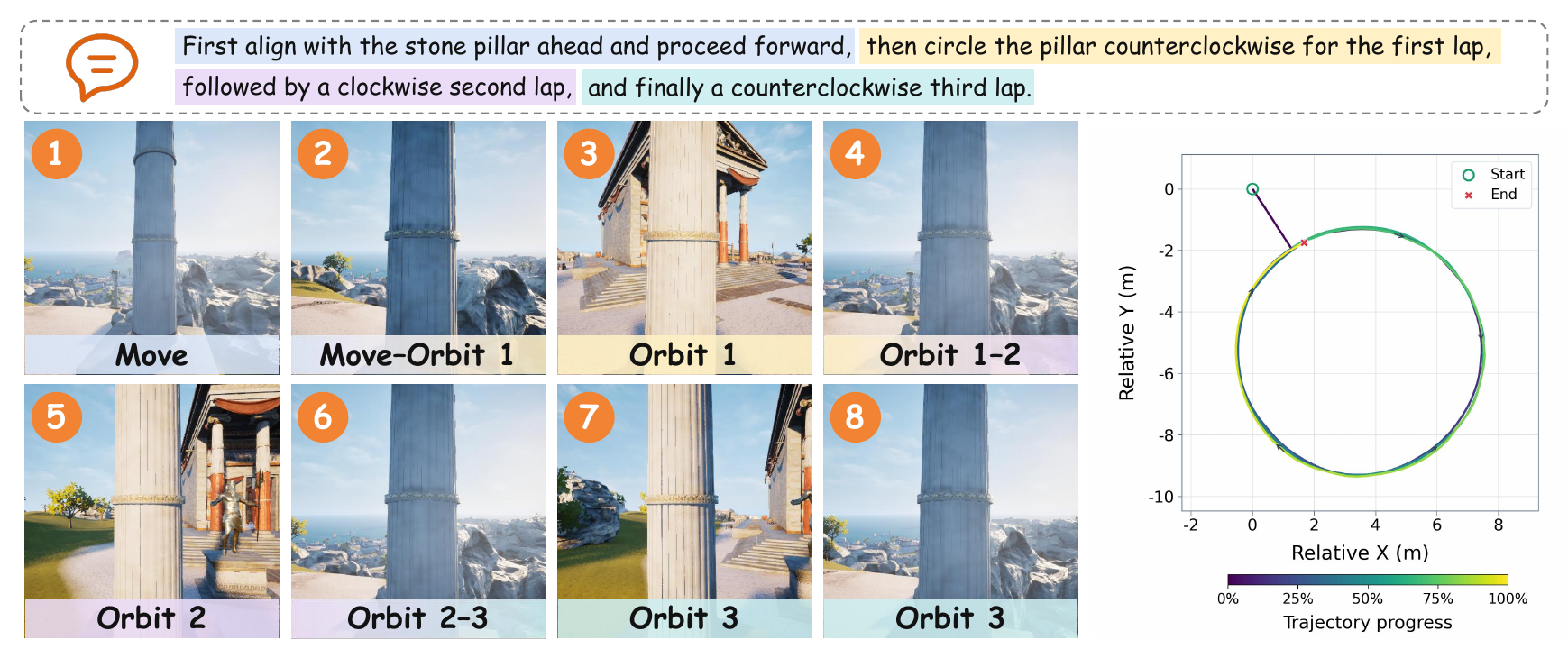}{\textbf{Move followed by three orbits.}
The UAV approaches the stone pillar and circles it counterclockwise, clockwise, and counterclockwise in successive laps.}

\casefigure{0.96}{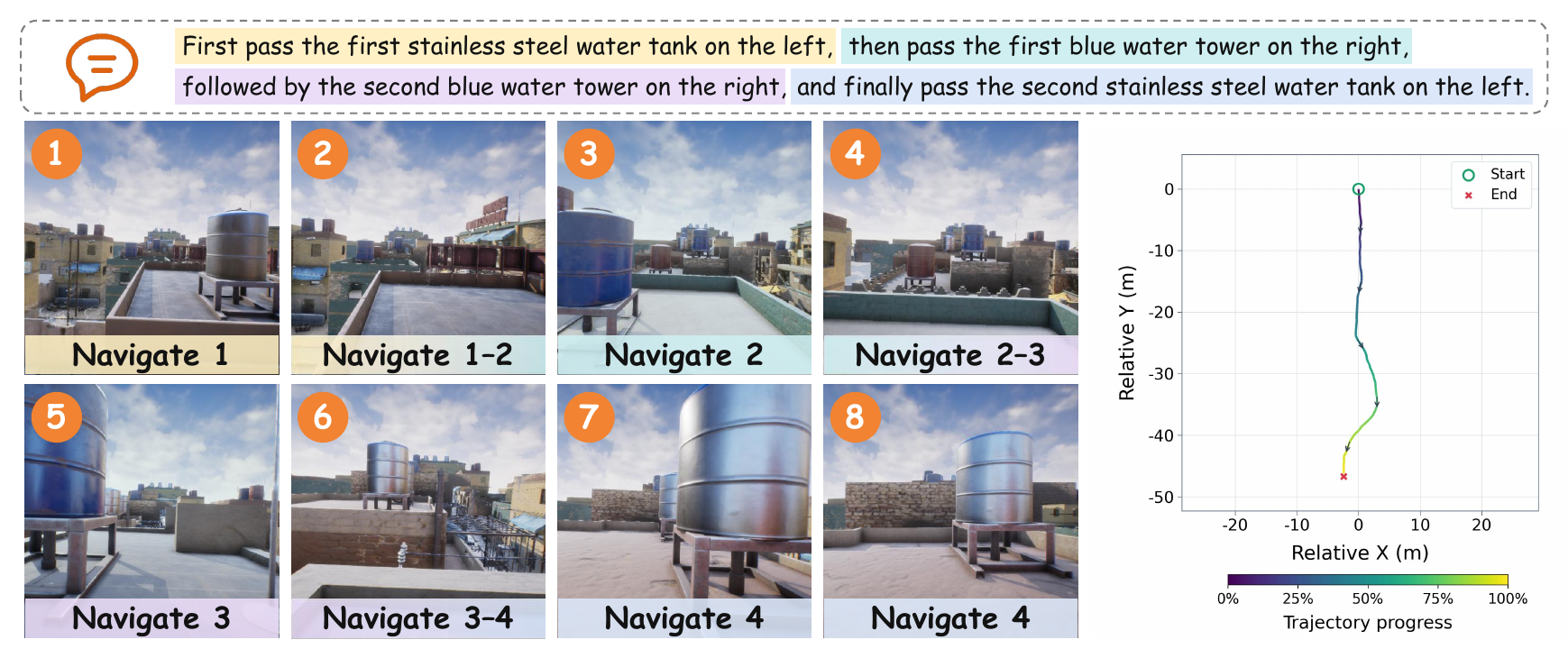}{\textbf{Four consecutive navigation skills.}
The UAV passes the first water tank on the left, both blue water towers on the right, and the second water tank on the left.}

\clearpage
\casefigure{0.88}{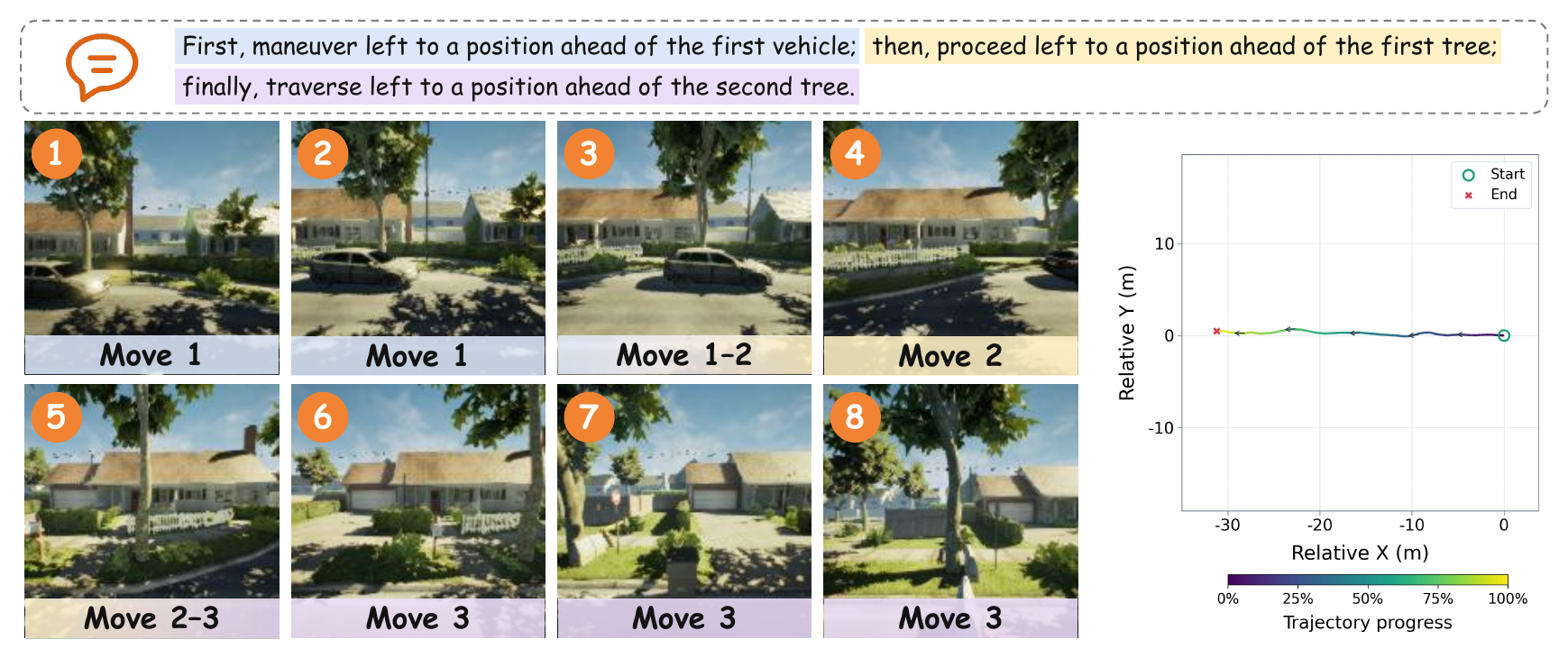}{\textbf{Three consecutive moves.}
The UAV moves left toward successive positions in front of a vehicle and two trees.}

\casefigure{0.88}{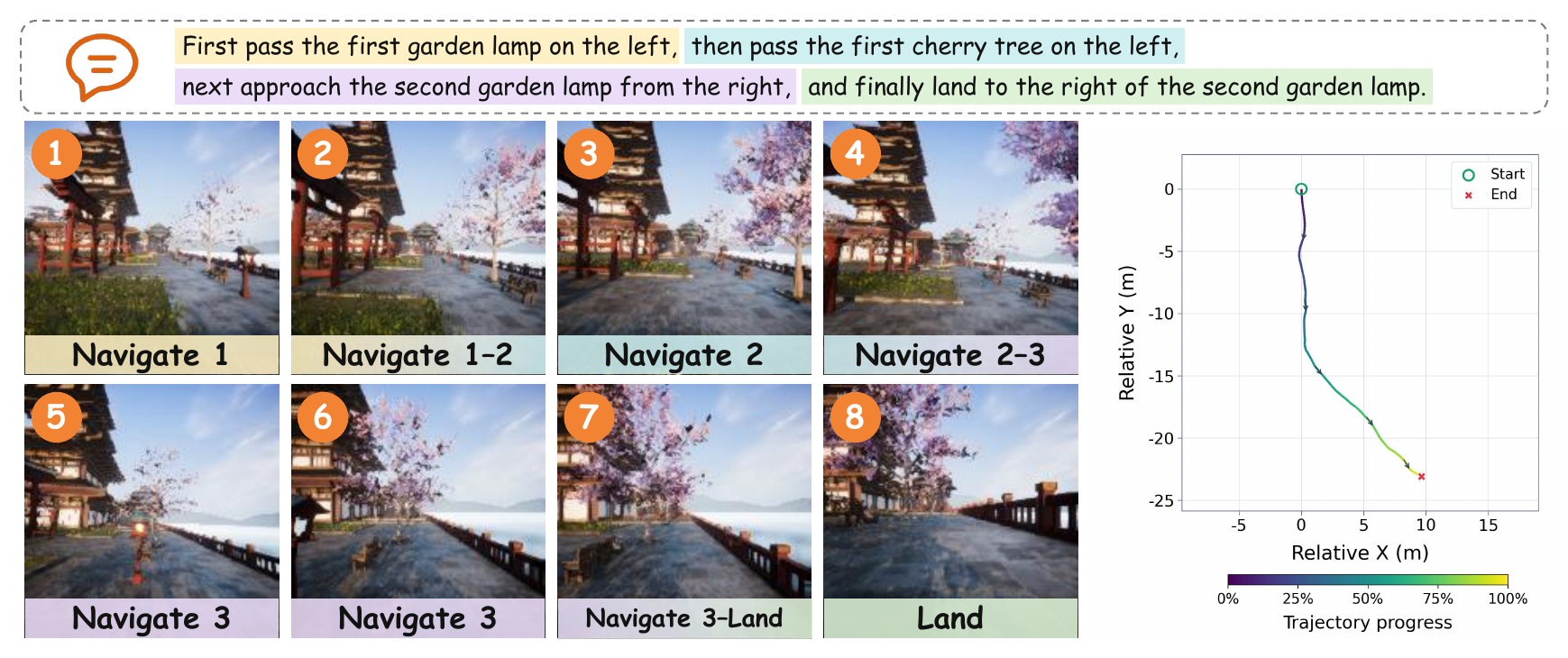}{\textbf{Navigation and landing near garden lamps.}
The UAV passes the first lamp and a cherry tree, approaches the second lamp from the right, and lands on its right.}

\casefigure{0.88}{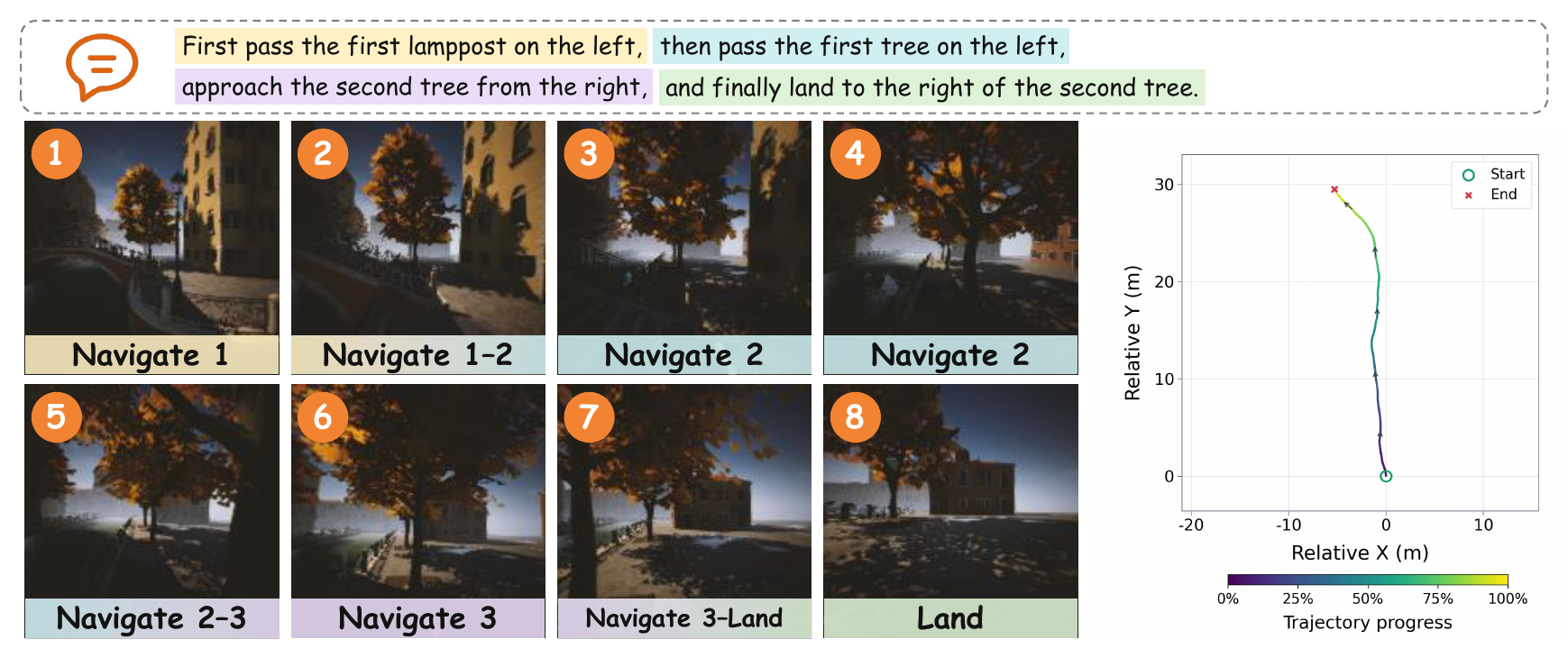}{\textbf{Navigation and landing near trees.}
The UAV passes a lamppost and the first tree, approaches the second tree from the right, and lands on its right.}

\clearpage
\subsection{Trajectory comparisons by perturbation timing}
\label{sec:qualitative-perturbation}
\noindent\begin{minipage}{\textwidth}
\centering
\includegraphics[width=0.74\textwidth]{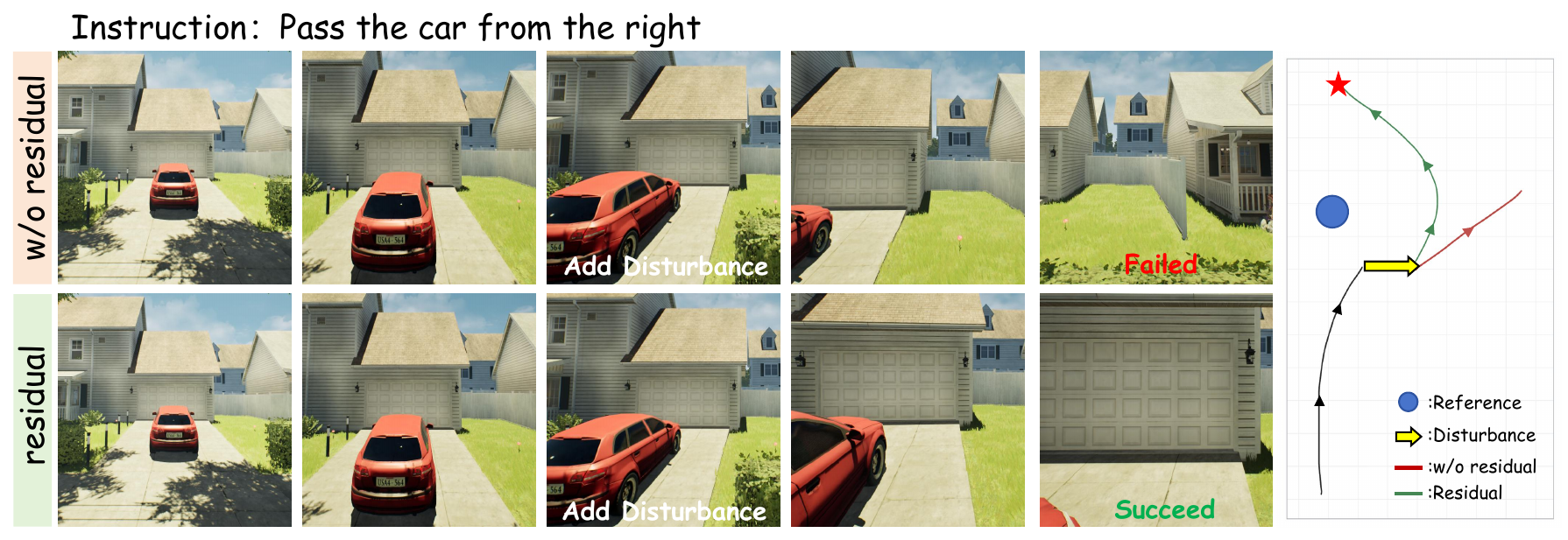}\par\smallskip
\includegraphics[width=0.74\textwidth]{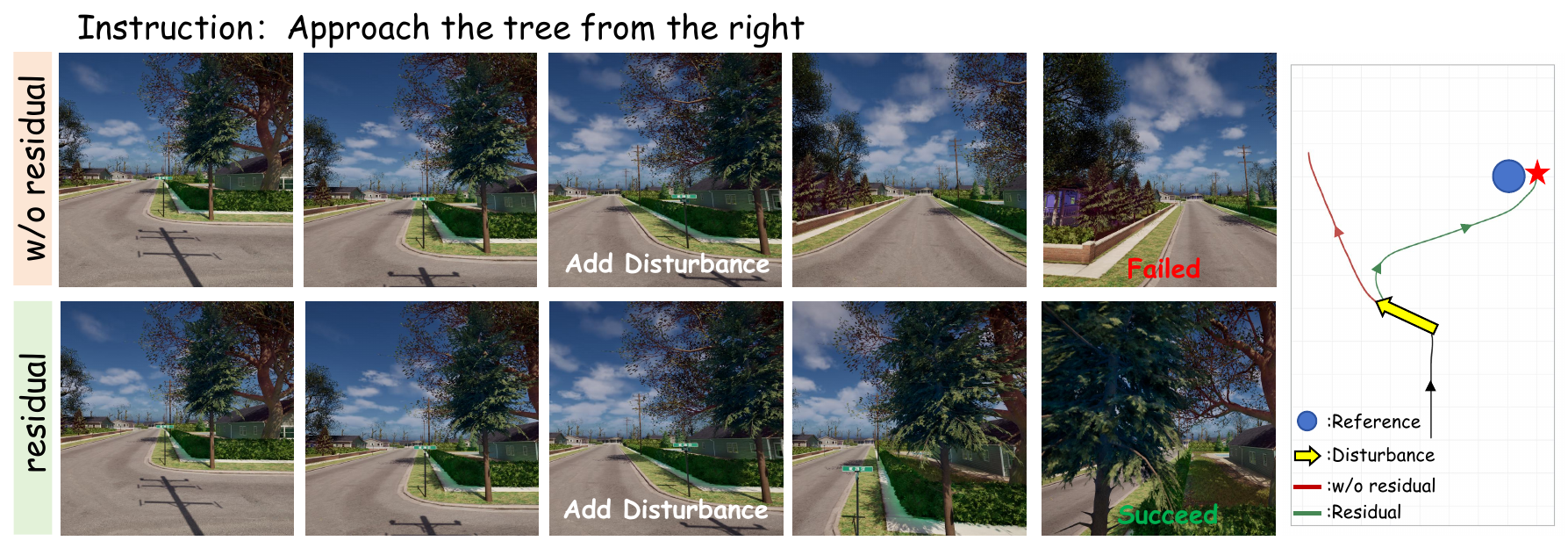}
\captionof{figure}{\textbf{Perturbation during skill execution.}
The UAV starts the skill and receives a pose perturbation partway through the maneuver. Both variants then continue from the disturbed state, with or without residual correction. The examples show a right-side car passage and a right-side approach to a tree.}
\end{minipage}

\medskip
\noindent\begin{minipage}{\textwidth}
\centering
\includegraphics[width=0.74\textwidth]{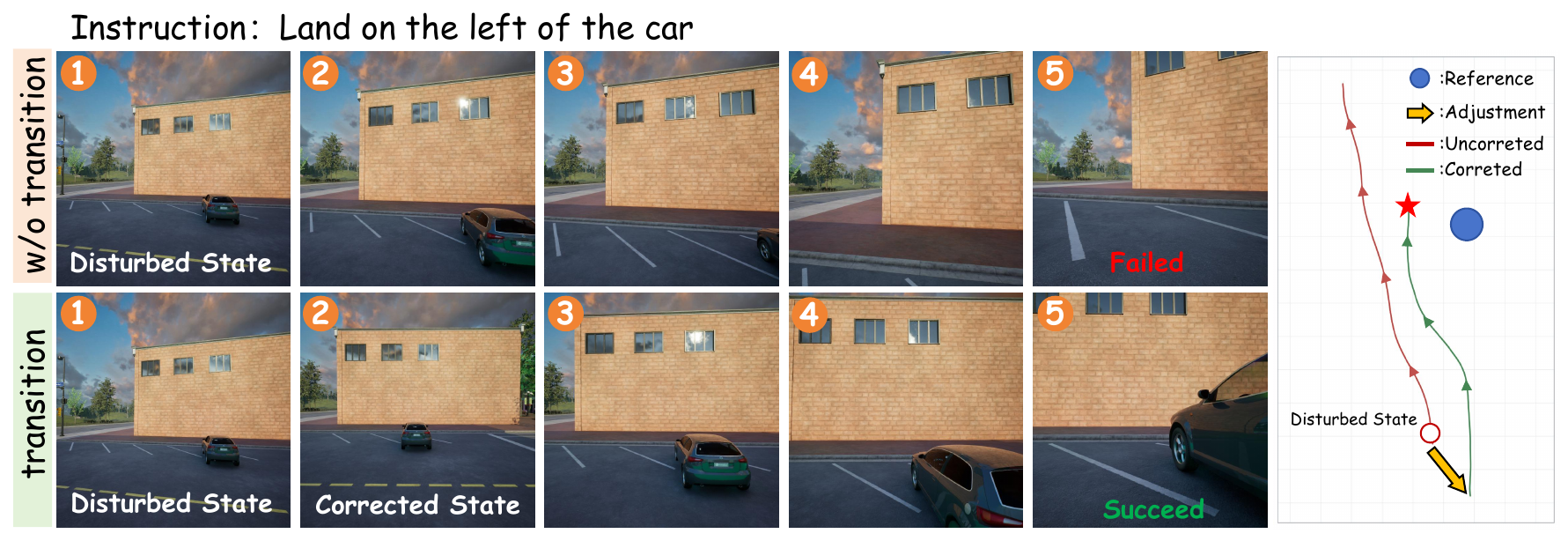}\par\smallskip
\includegraphics[width=0.74\textwidth]{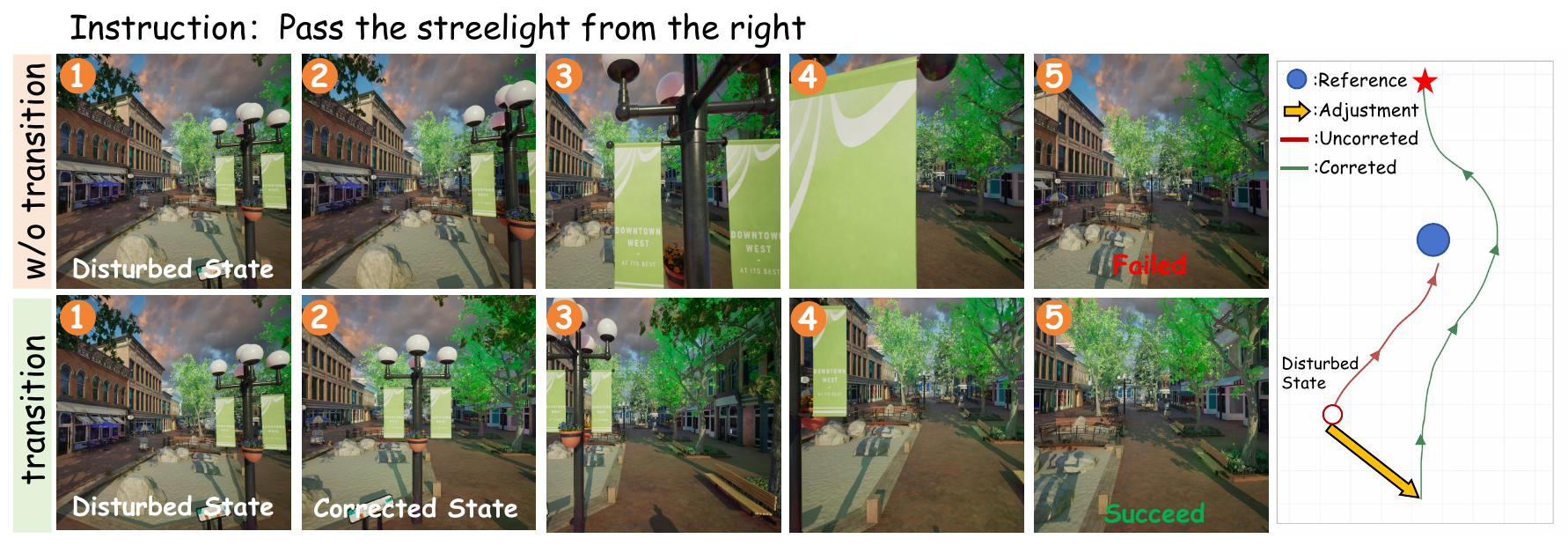}
\captionof{figure}{\textbf{Perturbation at the start of the skill.}
The pose is perturbed before the first action of the skill under evaluation. The transition-enabled variant applies one adjustment, and the other variant starts directly from the perturbed pose. The examples show a landing to the left of a car and a right-side streetlight passage.}
\end{minipage}

\clearpage
\section{Real-World Flight Examples}

We deploy AeroSkill on a FanciSwarm Pro UAV. An inference server with an NVIDIA RTX 6000 Ada GPU communicates with the aircraft over local Wi-Fi. AeroSkill parses each instruction once and executes the skills through the 5 Hz controller. Each method flies once on each of the same 100 instructions without deployment-specific fine-tuning. The keyframes on this page show two AeroSkill flights.

\casefigure{0.70}{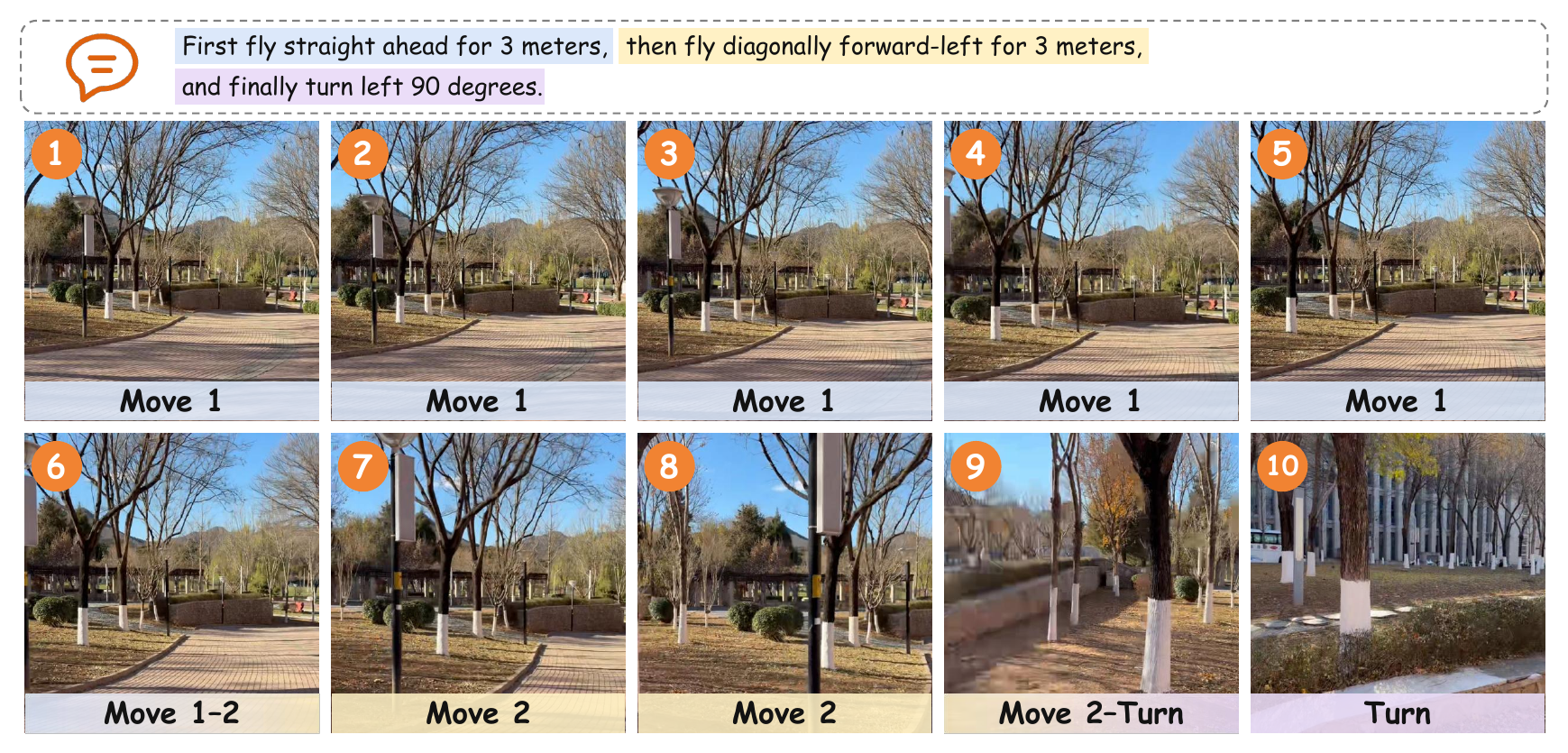}{\textbf{Two moves followed by a turn.}
The UAV flies forward for 3 m, moves diagonally forward-left for 3 m, and turns left by $90^\circ$.}

\casefigure{0.70}{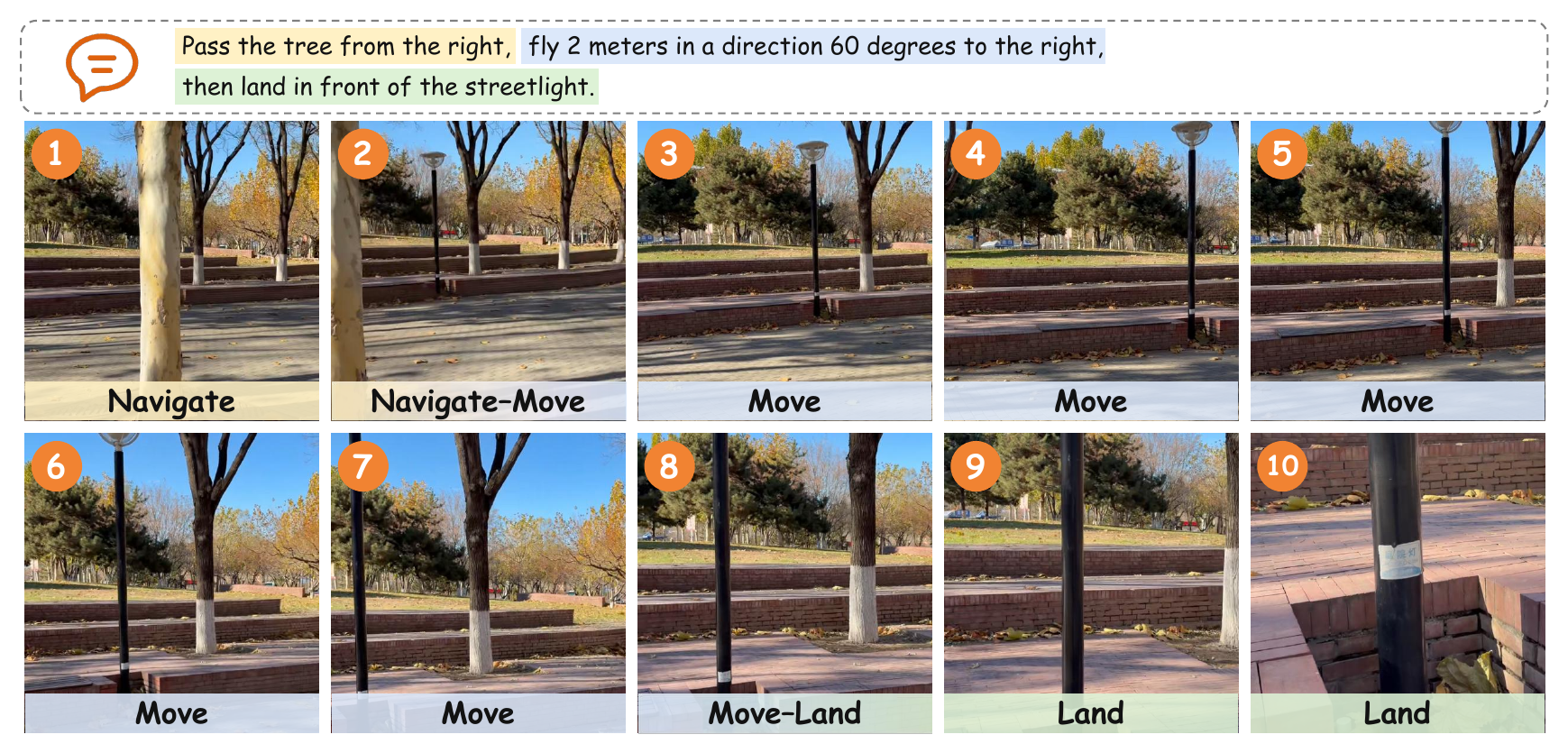}{\textbf{Tree passage followed by repositioning and landing.}
The UAV passes the tree on the right, flies 2 m in a direction $60^\circ$ to the right, and lands in front of the streetlight.}

\end{document}